\pdfoutput=1

\documentclass[11pt]{article}

\usepackage{acl}

\usepackage{times}
\usepackage{latexsym}
\usepackage{multirow}
\usepackage{booktabs}
\usepackage[T1]{fontenc}
\usepackage[utf8]{inputenc}
\usepackage{amsfonts}
\usepackage{amsmath}
\usepackage{amssymb}
\usepackage{microtype}
\usepackage{graphicx}
\usepackage{subcaption}
\usepackage{enumitem}
\usepackage{longtable}
\usepackage{xcolor} 
\usepackage{hyperref}
\usepackage{url}

\title{A Taxonomy of Programming Languages for Code Generation}

\author{Nishat Raihan$^{1}$, Christian Newman$^{2}$, Marcos Zampieri$^{1}$ \\
$^{1}$George Mason University, USA \\
$^{2}$Rochester Institute of Technology, USA \\
\texttt{mraihan2@gmu.edu}
}

\begin{document}
\maketitle

\begin{abstract}
The world's 7,000+ languages vary widely in the availability of resources for NLP, motivating efforts to systematically categorize them by their degree of resourcefulness \cite{joshi-etal-2020-state}. A similar disparity exists among programming languages (PLs); however, no resource-tier taxonomy has been established for code. As large language models (LLMs) grow increasingly capable of generating code, such a taxonomy becomes essential. To fill this gap, we present the first reproducible PL resource classification, grouping 646 languages into four tiers. We show that only 1.9\% of languages (Tier~3, High) account for 74.6\% of all tokens in seven major \textcolor{black}{web-derived} corpora, while 71.7\% of languages (Tier~0, Scarce) contribute just 1.0\%. Statistical analyses of within-tier inequality, dispersion, and distributional skew confirm that this imbalance is both extreme and systematic. Our results provide a principled framework for dataset curation and tier-aware evaluation of multilingual LLMs\textcolor{black}{, and a first quantitative picture of code availability on the web}.
\end{abstract}

\section{Introduction}

The advent of large language models trained on code (Code LLMs) has triggered a paradigm shift in software engineering \cite{chen2021evaluating,li2022competition,raihan2024code}. These models, powered by vast web-scraped corpora such as The Stack \cite{kocetkov2022stack} and CodeSearchNet \cite{husain2019codesearchnet}, can generate complex code, translate between programming languages, and accelerate software development. However, their capabilities are contingent on their training data, which is heavily skewed toward a few popular PLs like Python and JavaScript. This imbalance, noted in analyses of code corpora \cite{xu2022systematic}, creates a ``digital divide'' where resource-rich languages receive superior model support while less common languages lag behind, with potential downstream implications for both fairness and security \cite{fan2023automated, pearce2022asleep}.

While the existence of this disparity is acknowledged, the field lacks a systematic framework to quantify it. This challenge mirrors the well-documented problem of resource scarcity in NLP. The work by \citet{joshi-etal-2020-state} provides a data-driven classification of natural languages into resource tiers, becoming a foundational tool for measuring linguistic diversity and guiding efforts to support low-resource languages. We argue that the PL ecosystem requires a similar foundational effort to move from anecdotal observations to quantitative understanding, enabling rigorous evaluation of models on less-represented languages \cite{valdez2023can} and fostering development of truly multilingual code models \cite{nijkamp2023codegen2}.

To address this gap, we present the first large-scale classification of PLs based on their prevalence in the \textcolor{black}{web-derived} corpora that power modern Code LLMs. We collect, merge, and deduplicate seven prominent code datasets and quantify each language's presence via token counts\textcolor{black}{, a direct measure of availability on the web}. From this analysis, we introduce a four-tier classification scheme: \textit{high-}, \textit{medium-}, \textit{low-}, and \textit{scarce-resource}. This work makes three primary contributions:

\begin{enumerate}[label=\textbf{\texttt{C\arabic*}}]
    \item A comprehensive empirical analysis of PL distribution across modern code corpora.
    \item A clear, data-driven four-tier classification of resource availability for 646 PLs.
    \item A public release of our token counts and tier assignments to facilitate equitable model benchmarking, bias diagnosis, and targeted research into the long tail of PLs.
\end{enumerate}

\begin{figure*}[t!]
    \centering
    \includegraphics[width=.8\linewidth]{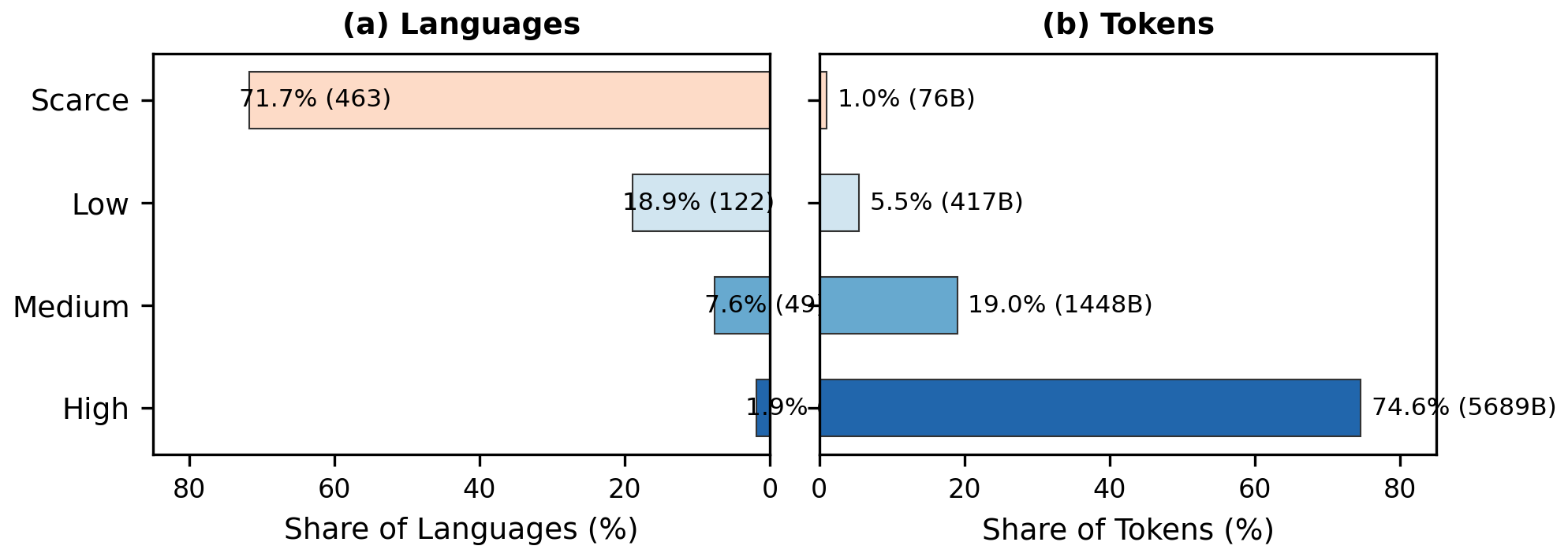}
    \caption{Resource imbalance across four tiers. \textbf{(a)}: share of languages; \textbf{(b)}: share of tokens. Only 1.9\% of languages (High) supply 74.6\% of all tokens, while 71.7\% of languages (Scarce) contribute just 1.0\%.}
    \label{fig:tier_imbalance}
\end{figure*}

\section{Related Work}
\label{sec:related}

\paragraph{Taxonomies for Natural Languages} The foundational work of \citet{joshi-etal-2020-state} introduced a six-class taxonomy for natural languages based on the availability of labeled and unlabeled data. Their framework has since guided data collection, model evaluation, and resource allocation across low-resource NLP. Our work adapts this philosophy to the PL domain, where data imbalance is equally pronounced but has lacked formal characterization.

\paragraph{Code Generation Data}
The rapid scaling of code pre-training data, from CodeSearchNet's six languages \cite{husain2019codesearchnet} to The Stack v2's 619 \cite{kocetkov2024stackv2}, has enabled increasingly capable Code LLMs \cite{chen2021evaluating, li2023starcoder, roziere2023codellama, guo2024deepseek}. However, the language composition of these corpora remains opaque. \citet{xu2022systematic} and \citet{raihan2024code} note that evaluations are overwhelmingly Python-centric, and recent surveys confirm that nearly 70\% of Code LLM studies involve Python-generated code \cite{raihan2025large}. Our taxonomy addresses this gap by providing a quantitative inventory of language representation across seven major corpora.

\paragraph{Multilingual Code Evaluation}
Benchmarks such as MultiPL-E \cite{cassano2023multipl}, HumanEval-X \cite{zheng2023codegeex}, mHumanEval \citet{raihan2024mhumaneval}, and the multilingual evaluation of \citet{athiwaratkun2023multilingual} have extended code evaluation beyond Python.  While these efforts are valuable, they remain limited to a small subset of ``popular'' PLs. Our taxonomy provides the missing link: a principled basis for selecting which languages to include and for interpreting performance differences through the lens of training-data availability.

\section{Methods}

\subsection{Data Collection}
\label{sec:data_sources}

Our analysis draws on seven corpora selected to represent the modern Code LLM data ecosystem (Table~\ref{tab:corpora}). The collection spans massive multilingual pre-training sets like The Stack v2 \cite{kocetkov2024stackv2} and RefineCode \cite{wang2024opencoder}, each covering over 600 languages with diverging curation philosophies: one based on exhaustive, graph-based collection and the other on aggressive filtering for safety and reproducibility.

Corpora with lower language counts serve specialized roles. StarCoder Data \cite{li2023starcoder} provides a curated 86-language subset. Project CodeNet \cite{puri2021project} offers rich problem-centric metadata for reasoning tasks, while CodeSearchNet \cite{husain2019codesearchnet} remains vital for code-retrieval models. The Heap \cite{katzy2025heap} is indispensable not as a training source but as a contamination-free evaluation benchmark. GitHub Code (CodeParrot) \cite{Touvron2023Llama2} provides a direct snapshot of public repository content. This diversity in scale, purpose, and curation philosophy provides a robust foundation for our classification.

\begin{table*}[!t]
\centering
\scalebox{0.8}{
\begin{tabular}{l l c c}
\toprule
\textbf{Corpus} & \textbf{Reported Size} & \textbf{\# of PLs} & \textbf{Reference} \\
\midrule
The Stack v2 &  900 B tokens & 619 & \citet{kocetkov2024stackv2} \\
RefineCode &  960 B tokens & 607 & \citet{wang2024opencoder} \\
StarCoder Data & \begin{tabular}[c]{@{}l@{}} 250 B tokens, 783 GB\end{tabular} & 86 & \citet{li2023starcoder} \\
The Heap &  32.7 M files & 57 & \citet{katzy2025heap} \\
Project CodeNet & \begin{tabular}[c]{@{}l@{}}14 M samples, 500 M LOC\end{tabular} & 55 & \citet{puri2021project} \\
GitHub Code (CodeParrot) & \begin{tabular}[c]{@{}l@{}}1 TB, 115 M files\end{tabular} & 32 & \citet{Touvron2023Llama2} \\
CodeSearchNet Corpus & \begin{tabular}[c]{@{}l@{}}20 GB, 6 M functions\end{tabular} & 6 & \citet{husain2019codesearchnet} \\
\bottomrule
\end{tabular}
}
\caption{Overview of the seven code corpora used in our study, sorted by number of programming languages (PLs). The selection represents the diversity of sources for training and evaluating modern Code LLMs.}
\label{tab:corpora}
\end{table*}

\subsection{Corpus Consolidation and Token Accounting}
\label{sec:validation}

To obtain reliable statistics, we fuse the seven raw corpora into a single cleaned dataset and measure it with a code-aware tokenizer. The workflow proceeds in three stages:

\begin{enumerate}[label=\textbf{\texttt{S\arabic*}}, leftmargin=*]
\item \textbf{Merge and Deduplicate.}
      We combine all corpora and drop exact duplicates by comparing SHA-256 hashes at file level, ensuring each unique source file contributes exactly once.
\item \textbf{PL Tagging.}
      We infer the programming language from the file extension using a refined map based on GitHub Linguist, which scales to millions of files while maintaining high precision across 600+ languages.
\item \textbf{Token Accounting.}
      We tokenize every file with the StarCoder tokenizer \cite{li2023starcoder}, record per-language token counts, and log auxiliary metrics (files, lines, bytes) for quality checks. Because the tokenizer mirrors how a competitive Code LLM ``sees'' code, these counts serve as a faithful proxy for model exposure.
\end{enumerate}

\section{Classification and Distribution}
\label{sec:classification}

For every one of the 646 programming languages $L$ in our corpus, we measure the total number of \emph{source-code tokens}, $t_L$.
Adapting the resource taxonomy for natural languages \cite{joshi-etal-2020-state}, we divide languages into four tiers:
\begin{align*}
\small
\text{Tier}_3 &: t_L \ge 100\text{B}; \\
\small
\text{Tier}_2 &: 10\text{B} \le t_L < 100\text{B}; \\
\small
\text{Tier}_1 &: 1\text{B} \le t_L < 10\text{B}; \\
\small
\text{Tier}_0 &: t_L < 1\text{B}.
\end{align*}

\noindent The corpus shows a pronounced long tail (Figure~\ref{fig:tier_imbalance}): Tier~0 covers 71.7\% of the 646 languages yet supplies only 1.0\% of the 7.63T tokens, whereas the 12 Tier~3 languages (1.9\%) supply 74.6\%.  Table~\ref{tab:resource_classes} summarizes the four tiers of taxonomy. 

\begin{table}[!ht]
\centering
\scalebox{0.75}{
\begin{tabular}{l c c l}
\toprule
\textbf{Tier} & \textbf{\# PLs} & \textbf{Tokens (B)} & \textbf{Notable Examples} \\
\midrule
High   & 12  & 5,689 & Python, JS, Java, C\#, C++ \\
Medium & 49  & 1,448 & SQL, Go, Kotlin, Rust, R \\
Low    & 122 & 417  & Prolog, Fortran, Julia, OCaml \\
Scarce & 463 & 76   & Ada, COBOL, Awk, GAMS \\
\bottomrule
\end{tabular}
}
\caption{Four-tier resource classification. Only 1.9\% of languages (High) account for 74.6\% of all 7.63T tokens; 71.7\% of languages (Scarce) contribute just 1.0\%.}
\label{tab:resource_classes}
\end{table}

\noindent Table~\ref{tab:tb10} lists the ten most-resourced languages and the ten least-resourced languages studied. The top-10 languages together hold 71.6\% of all tokens. Benchmarks dominated by these languages risk overstating model robustness \cite{cassano2023multipl,zheng2023codegeex,athiwaratkun2023multilingual}. On the other hand, the ten least-resourced languages together account for only 0.00003\% of tokens (roughly 2.02M of 7.63T). Despite their rarity, these niche formats and DSLs (e.g., templates, tracebacks, meta-languages) are vital for real-world tooling and software analysis and merit targeted data collection. The full taxonomy is presented in Appendix~\ref{app:classification}.

\begin{table}[!ht]
\centering
\scalebox{0.78}{
\begin{tabular}{r l c l}
\toprule
\textbf{Rank} & \textbf{Language} & \textbf{Tier} & \textbf{Resource}\\
\midrule
1  & Python      & 3 & High\\
2  & JavaScript  & 3 & High\\
3  & Java        & 3 & High\\
4  & C\#         & 3 & High\\
5  & PHP         & 3 & High\\
6  & C++         & 3 & High\\
7  & CSS         & 3 & High\\
8  & TypeScript  & 3 & High\\
9  & C           & 3 & High\\
10 & Ruby        & 3 & High\\
\midrule
637 & Myghty                        & 0 & Scarce\\
638 & Object Data Instance Notation & 0 & Scarce\\
639 & Ecere Projects                & 0 & Scarce\\
640 & NumPy                         & 0 & Scarce\\
641 & Omgrofl                       & 0 & Scarce\\
642 & Python traceback              & 0 & Scarce\\
643 & MiniD                         & 0 & Scarce\\
644 & NetLinx+ERB                   & 0 & Scarce\\
645 & Record Jar                    & 0 & Scarce\\
646 & C-ObjDump                     & 0 & Scarce\\
\bottomrule
\end{tabular}}
\caption{Top and bottom-10 programming languages.}
\label{tab:tb10}
\end{table}

\begin{figure*}[!ht]
    \centering
    \includegraphics[width=.8\linewidth]{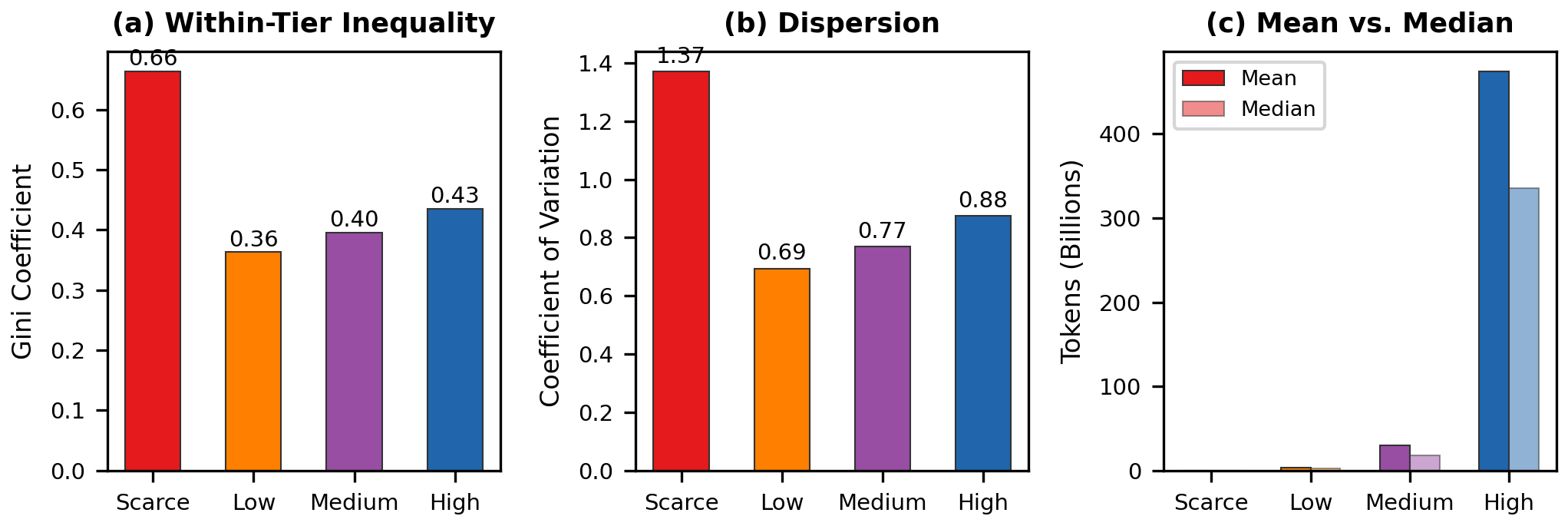}
    \caption{Within-tier statistics: \textbf{(a)} Gini coefficient measuring token inequality; \textbf{(b)} coefficient of variation measuring dispersion; \textbf{(c)} mean vs.\ median tokens per language, highlighting right-skew in every tier.}
    \label{fig:stats_panel}
\end{figure*}

\section{Significance and Analysis}
\label{sec:significance}

State-of-the-art Code LLMs, such as StarCoder \cite{li2023starcoder} and Code Llama \cite{roziere2023codellama}, train on multi-terabyte corpora whose language mix is opaque and heavily skewed toward a handful of ``major'' PLs. Inspired by the NLP resource taxonomy of \citet{joshi-etal-2020-state}, we introduce a transparent, token-count-based classification that \emph{systematically} exposes and quantifies this skew, rather than relying on ad hoc labels.

\paragraph{Imbalances}
The Gini coefficient (Figure~\ref{fig:stats_panel}a) reveals that inequality is most severe within the Scarce tier (Gini\,=\,0.66), where a few languages near the 1B boundary dwarf hundreds of near-zero entries. The High tier also shows concentration (Gini\,=\,0.43), since Python alone holds ${\sim}$27\% of Tier~3 tokens. Without such diagnostics, aggregate metrics would be dominated by Python and C++ while dozens of low-resource PLs remain invisible \cite{muennighoff2023crosslingual, raihan2024mhumaneval}.

\paragraph{Variation}
The coefficient of variation (Figure~\ref{fig:stats_panel}b) identifies which tiers contain the most heterogeneous PL populations. The Scarce tier's CV of 1.37 signals that targeted scraping, augmentation, or synthetic generation could disproportionately benefit the most under-served languages. The mean-vs.-median comparison (Figure~\ref{fig:stats_panel}c) further highlights the right-skewed nature of every tier: means consistently exceed medians, confirming that a few dominant languages inflate tier-level averages.

\begin{figure}[!t]
    \centering
    \includegraphics[width=0.75\linewidth]{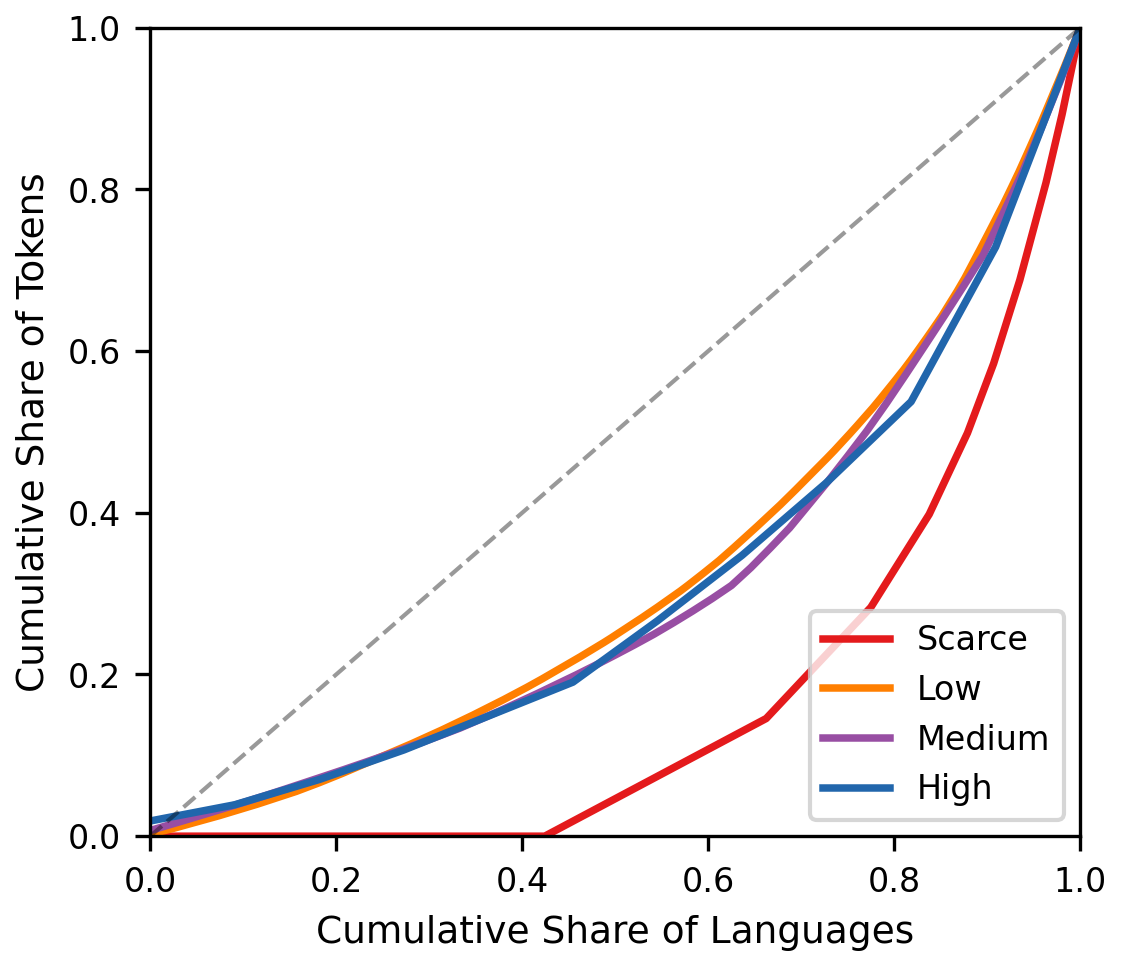}
    \caption{Curves of token distribution within each tier; deviation from the diagonal indicates concentration.}
    \label{fig:lorenz_tier}
\end{figure}

\begin{figure}[!t]
    \centering
    \includegraphics[width=.75\linewidth]{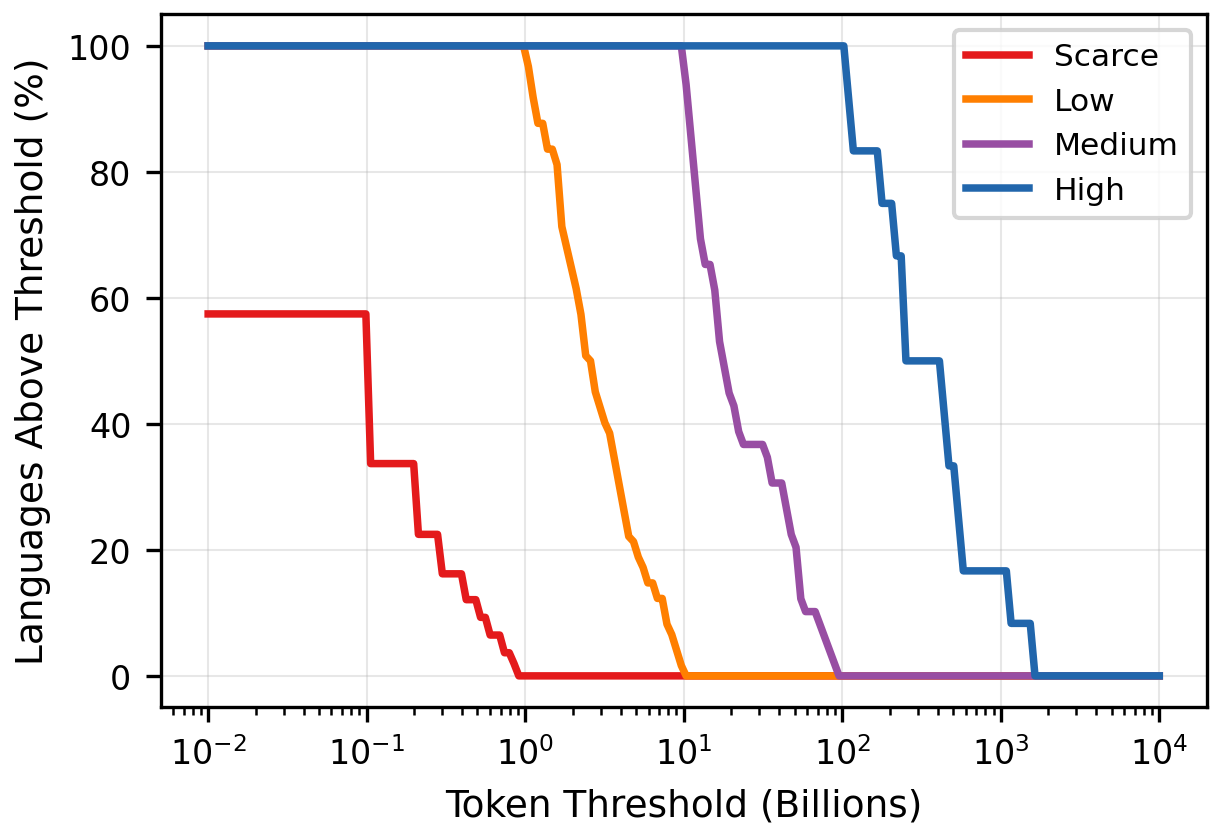}
    \caption{Percentage of languages above token thresholds (log scale), comparing survivability across tiers.}
    \label{fig:ecdf_tier}
\end{figure}

\paragraph{Tier-aware Evaluation.}
The Lorenz curves (Figure~\ref{fig:lorenz_tier}) show that deviation from perfect equality is most pronounced in the High and Medium tiers, indicating that even within resource-rich 
groupings, token allocation is far from uniform. The ECDF survival curves (Figure~\ref{fig:ecdf_tier}) complement this view: at a threshold of 10B tokens, 100\% of High-tier languages survive but only ${\sim}$45\% of Medium-tier languages do, with Low and Scarce tiers dropping sharply below 1B. These visualizations motivate stratified scoring in benchmarks and leaderboards, rather than relying on overall means that mask per-tier weaknesses.

\paragraph{Infrastructure}
As corpora like The Stack v2 \cite{kocetkov2024stackv2} continue to grow, the taxonomy offers a reproducible scaffold to diagnose bias and track progress. It also provides a lens for interpreting model behavior: when a Code LLM defaults to Python under a language-neutral prompt, the taxonomy quantifies \emph{why}, since Python commands 1,543B tokens (20.2\% of the corpus), dwarfing all other languages. By linking training-data composition to generation bias, the taxonomy transforms ``Python default'' from an anecdote into a measurable, tier-level phenomenon.

\section{Conclusion}

We present the first tiered, token-count-based taxonomy to categorize PL resource levels. Our analysis of seven major \textcolor{black}{web-derived} code corpora covering 646 PLs and 7.63 trillion tokens reveals stark imbalance: 12 High-resource languages hold 74.6\% of all tokens, while 463 Scarce-resource languages share just 1.0\%. Gini coefficients, Lorenz curves, and CV analysis confirm that this imbalance persists both across and within tiers.

Existing multilingual benchmarks \cite{cassano2023multipl, zheng2023codegeex, athiwaratkun2023multilingual, raihan2024mhumaneval} evaluate beyond Python but select languages without accounting for training-data availability. Our taxonomy is the first framework to not only identify under-resourced PLs but quantify the magnitude of that scarcity. \textcolor{black}{Viewed through a web-as-corpus lens, it also documents which PLs the web over- and under-represents, guiding future crawling and curation efforts.}

\section*{Limitations}

Our choices of seven permissively licensed corpora, a single StarCoder tokenizer, extension-based PL detection, exact-hash deduplication, and heuristic tier cutoffs are intentional for transparency, reproducibility, and alignment with how current Code LLMs are actually trained. These design points may influence absolute counts (e.g., polyglot files, boilerplate variants), but they do not disturb the heavy-tailed shape or the rank-order skew our taxonomy exposes. We release raw counts so anyone can swap tokenizers, tweak thresholds, or apply finer-grained deduplication without rerunning the entire pipeline. In short, these are controlled methodological knobs: they might refine decimal values, but they do not overturn the central finding that code resources are starkly imbalanced and demand systematic accounting.

\section*{Ethical Considerations}

We follow the ACL Code of Ethics and use only permissively licensed, publicly accessible data, respecting original licenses and platform terms of service. No proprietary or personal data were intentionally included, and we release only aggregate statistics, not raw code, to minimize privacy and licensing risks.

\bibliography{custom}

\clearpage

\appendix

\onecolumn
\section{The Full PL Classification}
\label{app:classification}

\small
\begin{center}
\begin{longtable}{@{}rlrrrl@{}}

\toprule
Rank & Language & File Count & Tokens (Billions) & Resource\_Tier & Tier\_Name \\
\midrule
\endfirsthead

\toprule
Rank & Language & File Count & Tokens (Billions) & Resource\_Tier & Tier\_Name \\
\midrule
\endhead

\midrule
\multicolumn{6}{c}{\textit{Continued on next page}}\\
\midrule
\endfoot

\bottomrule
\endlastfoot
1 & Python & 154281537 & 1542.8 & 3 & High \\
2 & JavaScript & 108871932 & 1088.7 & 3 & High \\
3 & Java & 56933829 & 569.3 & 3 & High \\
4 & C\# & 51233056 & 512.3 & 3 & High \\
5 & PHP & 46025156 & 460.3 & 3 & High \\
6 & C++ & 43175936 & 431.8 & 3 & High \\
7 & CSS & 23869063 & 238.7 & 3 & High \\
8 & TypeScript & 23853979 & 238.5 & 3 & High \\
9 & C & 20775502 & 207.8 & 3 & High \\
10 & Ruby & 17785998 & 177.9 & 3 & High \\
11 & Objective-C & 11407073 & 114.1 & 3 & High \\
12 & Shell & 10681876 & 106.8 & 3 & High \\
13 & Go & 9301819 & 93.0 & 2 & Medium \\
14 & Kotlin & 8616859 & 86.2 & 2 & Medium \\
15 & Jupyter Notebook & 8151402 & 81.5 & 2 & Medium \\
16 & Vue & 7375658 & 73.8 & 2 & Medium \\
17 & Swift & 7234581 & 72.3 & 2 & Medium \\
18 & TSX & 5496011 & 55.0 & 2 & Medium \\
19 & Java Server Pages & 5394783 & 53.9 & 2 & Medium \\
20 & SQL & 5285705 & 52.9 & 2 & Medium \\
21 & INI & 5261835 & 52.6 & 2 & Medium \\
22 & R & 5145606 & 51.5 & 2 & Medium \\
23 & Blade & 4925151 & 49.3 & 2 & Medium \\
24 & MATLAB & 4543828 & 45.4 & 2 & Medium \\
25 & Scala & 4445036 & 44.5 & 2 & Medium \\
26 & Makefile & 4221531 & 42.2 & 2 & Medium \\
27 & Dart & 4214307 & 42.1 & 2 & Medium \\
28 & Java Properties & 3611528 & 36.1 & 2 & Medium \\
29 & HTML+Razor & 3510527 & 35.1 & 2 & Medium \\
30 & TeX & 3190550 & 31.9 & 2 & Medium \\
31 & Lua & 2347887 & 23.5 & 2 & Medium \\
32 & Rust & 2218033 & 22.2 & 2 & Medium \\
33 & TSV & 2116969 & 21.2 & 2 & Medium \\
34 & Smali & 2039732 & 20.4 & 2 & Medium \\
35 & Dockerfile & 1898900 & 19.0 & 2 & Medium \\
36 & reStructuredText & 1826351 & 18.3 & 2 & Medium \\
37 & Hack & 1786841 & 17.9 & 2 & Medium \\
38 & CMake & 1743194 & 17.4 & 2 & Medium \\
39 & Less & 1666356 & 16.7 & 2 & Medium \\
40 & Microsoft Visual Studio Solution & 1665208 & 16.7 & 2 & Medium \\
41 & WebVTT & 1603735 & 16.0 & 2 & Medium \\
42 & Gradle & 1598251 & 16.0 & 2 & Medium \\
43 & Pug & 1503380 & 15.0 & 2 & Medium \\
44 & EJS & 1497213 & 15.0 & 2 & Medium \\
45 & Twig & 1316871 & 13.2 & 2 & Medium \\
46 & ASP.NET & 1287527 & 12.9 & 2 & Medium \\
47 & Haskell & 1254071 & 12.5 & 2 & Medium \\
48 & Diff & 1204035 & 12.0 & 2 & Medium \\
49 & Gherkin & 1203914 & 12.0 & 2 & Medium \\
50 & Handlebars & 1188119 & 11.9 & 2 & Medium \\
51 & TSQL & 1155064 & 11.6 & 2 & Medium \\
52 & Wikitext & 1147269 & 11.5 & 2 & Medium \\
53 & Perl & 1146878 & 11.5 & 2 & Medium \\
54 & Batchfile & 1127069 & 11.3 & 2 & Medium \\
55 & Groovy & 1103831 & 11.0 & 2 & Medium \\
56 & Public Key & 1062027 & 10.6 & 2 & Medium \\
57 & Visual Basic .NET & 1060179 & 10.6 & 2 & Medium \\
58 & Haml & 1046278 & 10.5 & 2 & Medium \\
59 & CoffeeScript & 1027516 & 10.3 & 2 & Medium \\
60 & Smalltalk & 1025689 & 10.3 & 2 & Medium \\
61 & ActionScript & 1013150 & 10.1 & 2 & Medium \\
62 & RMarkdown & 994233 & 9.9 & 1 & Low \\
63 & Wavefront Object & 984476 & 9.8 & 1 & Low \\
64 & Pascal & 951936 & 9.5 & 1 & Low \\
65 & PostScript & 950368 & 9.5 & 1 & Low \\
66 & Roff & 936321 & 9.4 & 1 & Low \\
67 & Org & 893315 & 8.9 & 1 & Low \\
68 & Gerber Image & 887319 & 8.9 & 1 & Low \\
69 & Elixir & 881706 & 8.8 & 1 & Low \\
70 & Checksums & 833913 & 8.3 & 1 & Low \\
71 & TOML & 815905 & 8.2 & 1 & Low \\
72 & Assembly & 773883 & 7.7 & 1 & Low \\
73 & Smarty & 769939 & 7.7 & 1 & Low \\
74 & Sass & 756458 & 7.6 & 1 & Low \\
75 & Maven POM & 749908 & 7.5 & 1 & Low \\
76 & HCL & 749763 & 7.5 & 1 & Low \\
77 & PowerShell & 682789 & 6.8 & 1 & Low \\
78 & Pickle & 643341 & 6.4 & 1 & Low \\
79 & Verilog & 636016 & 6.4 & 1 & Low \\
80 & Processing & 578171 & 5.8 & 1 & Low \\
81 & Vim Script & 572386 & 5.7 & 1 & Low \\
82 & mIRC Script & 561872 & 5.6 & 1 & Low \\
83 & GLSL & 548969 & 5.5 & 1 & Low \\
84 & Roff Manpage & 525533 & 5.3 & 1 & Low \\
85 & Game Maker Language & 500777 & 5.0 & 1 & Low \\
86 & Unix Assembly & 494749 & 4.9 & 1 & Low \\
87 & Gettext Catalog & 490963 & 4.9 & 1 & Low \\
88 & HTML+PHP & 478802 & 4.8 & 1 & Low \\
89 & VHDL & 451684 & 4.5 & 1 & Low \\
90 & Julia & 447047 & 4.5 & 1 & Low \\
91 & Motorola 68K Assembly & 442526 & 4.4 & 1 & Low \\
92 & Graphviz (DOT) & 433493 & 4.3 & 1 & Low \\
93 & Scheme & 426836 & 4.3 & 1 & Low \\
94 & Apex & 408075 & 4.1 & 1 & Low \\
95 & OpenEdge ABL & 403912 & 4.0 & 1 & Low \\
96 & Go Module & 398105 & 4.0 & 1 & Low \\
97 & AsciiDoc & 396165 & 4.0 & 1 & Low \\
98 & Stylus & 395048 & 4.0 & 1 & Low \\
99 & CoNLL-U & 393103 & 3.9 & 1 & Low \\
100 & Clojure & 389392 & 3.9 & 1 & Low \\
101 & Slim & 367031 & 3.7 & 1 & Low \\
102 & Haxe & 366379 & 3.7 & 1 & Low \\
103 & Emacs Lisp & 365757 & 3.7 & 1 & Low \\
104 & OCaml & 364452 & 3.6 & 1 & Low \\
105 & SubRip Text & 362904 & 3.6 & 1 & Low \\
106 & Git Config & 360341 & 3.6 & 1 & Low \\
107 & Fortran Free Form & 352647 & 3.5 & 1 & Low \\
108 & AMPL & 352293 & 3.5 & 1 & Low \\
109 & Jest Snapshot & 318704 & 3.2 & 1 & Low \\
110 & Common Lisp & 315454 & 3.2 & 1 & Low \\
111 & BibTeX & 314691 & 3.1 & 1 & Low \\
112 & Rich Text Format & 305866 & 3.1 & 1 & Low \\
113 & Erlang & 303085 & 3.0 & 1 & Low \\
114 & GDScript & 284996 & 2.8 & 1 & Low \\
115 & Jinja & 278431 & 2.8 & 1 & Low \\
116 & JSON with Comments & 277494 & 2.8 & 1 & Low \\
117 & Fortran & 271017 & 2.7 & 1 & Low \\
118 & F\# & 266595 & 2.7 & 1 & Low \\
119 & Fluent & 264102 & 2.6 & 1 & Low \\
120 & Puppet & 255916 & 2.6 & 1 & Low \\
121 & XML Property List & 255867 & 2.6 & 1 & Low \\
122 & JAR Manifest & 255357 & 2.6 & 1 & Low \\
123 & D & 253920 & 2.5 & 1 & Low \\
124 & VBA & 244935 & 2.4 & 1 & Low \\
125 & edn & 240030 & 2.4 & 1 & Low \\
126 & desktop & 239371 & 2.4 & 1 & Low \\
127 & Svelte & 234670 & 2.3 & 1 & Low \\
128 & Gnuplot & 234425 & 2.3 & 1 & Low \\
129 & Protocol Buffer & 234297 & 2.3 & 1 & Low \\
130 & PLSQL & 231741 & 2.3 & 1 & Low \\
131 & Nix & 230308 & 2.3 & 1 & Low \\
132 & M & 223873 & 2.2 & 1 & Low \\
133 & PureBasic & 220170 & 2.2 & 1 & Low \\
134 & Turtle & 213057 & 2.1 & 1 & Low \\
135 & Raku & 211082 & 2.1 & 1 & Low \\
136 & Raw token data & 207118 & 2.1 & 1 & Low \\
137 & Tcl & 204437 & 2.0 & 1 & Low \\
138 & mcfunction & 201109 & 2.0 & 1 & Low \\
139 & JavaScript+ERB & 198008 & 2.0 & 1 & Low \\
140 & Racket & 197694 & 2.0 & 1 & Low \\
141 & Classic ASP & 193001 & 1.9 & 1 & Low \\
142 & Prolog & 187485 & 1.9 & 1 & Low \\
143 & QMake & 187415 & 1.9 & 1 & Low \\
144 & Gentoo Ebuild & 185598 & 1.9 & 1 & Low \\
145 & PLpgSQL & 183383 & 1.8 & 1 & Low \\
146 & QML & 182471 & 1.8 & 1 & Low \\
147 & Solidity & 180287 & 1.8 & 1 & Low \\
148 & DIGITAL Command Language & 179931 & 1.8 & 1 & Low \\
149 & SQF & 173685 & 1.7 & 1 & Low \\
150 & Elm & 173109 & 1.7 & 1 & Low \\
151 & SystemVerilog & 171959 & 1.7 & 1 & Low \\
152 & Cuda & 171244 & 1.7 & 1 & Low \\
153 & Procfile & 170824 & 1.7 & 1 & Low \\
154 & KiCad Layout & 167029 & 1.7 & 1 & Low \\
155 & Mathematica & 166560 & 1.7 & 1 & Low \\
156 & RPM Spec & 165787 & 1.7 & 1 & Low \\
157 & XSLT & 165580 & 1.7 & 1 & Low \\
158 & Standard ML & 164588 & 1.6 & 1 & Low \\
159 & GraphQL & 160523 & 1.6 & 1 & Low \\
160 & Groovy Server Pages & 157685 & 1.6 & 1 & Low \\
161 & E-mail & 149672 & 1.5 & 1 & Low \\
162 & ApacheConf & 148905 & 1.5 & 1 & Low \\
163 & COLLADA & 146726 & 1.5 & 1 & Low \\
164 & Starlark & 134350 & 1.3 & 1 & Low \\
165 & Objective-C++ & 133114 & 1.3 & 1 & Low \\
166 & Scilab & 132876 & 1.3 & 1 & Low \\
167 & Liquid & 130642 & 1.3 & 1 & Low \\
168 & PlantUML & 129291 & 1.3 & 1 & Low \\
169 & IDL & 122615 & 1.2 & 1 & Low \\
170 & Wavefront Material & 121616 & 1.2 & 1 & Low \\
171 & Stata & 121052 & 1.2 & 1 & Low \\
172 & STL & 119027 & 1.2 & 1 & Low \\
173 & fish & 116039 & 1.2 & 1 & Low \\
174 & LLVM & 113925 & 1.1 & 1 & Low \\
175 & HLSL & 113570 & 1.1 & 1 & Low \\
176 & xBase & 113188 & 1.1 & 1 & Low \\
177 & Forth & 109194 & 1.1 & 1 & Low \\
178 & Mustache & 109049 & 1.1 & 1 & Low \\
179 & ColdFusion & 105315 & 1.1 & 1 & Low \\
180 & FreeMarker & 104830 & 1.0 & 1 & Low \\
181 & Crystal & 102607 & 1.0 & 1 & Low \\
182 & Nunjucks & 100455 & 1.0 & 1 & Low \\
183 & Git Attributes & 97737 & 1.0 & 1 & Low \\
184 & AppleScript & 94262 & 0.9 & 0 & Scarce \\
185 & 1C Enterprise & 94066 & 0.9 & 0 & Scarce \\
186 & Coq & 92375 & 0.9 & 0 & Scarce \\
187 & YASnippet & 92294 & 0.9 & 0 & Scarce \\
188 & Ada & 92104 & 0.9 & 0 & Scarce \\
189 & X PixMap & 90219 & 0.9 & 0 & Scarce \\
190 & Web Ontology Language & 89435 & 0.9 & 0 & Scarce \\
191 & LilyPond & 87374 & 0.9 & 0 & Scarce \\
192 & robots.txt & 85086 & 0.9 & 0 & Scarce \\
193 & SaltStack & 82776 & 0.8 & 0 & Scarce \\
194 & Ant Build System & 82227 & 0.8 & 0 & Scarce \\
195 & Nim & 81595 & 0.8 & 0 & Scarce \\
196 & ReScript & 80682 & 0.8 & 0 & Scarce \\
197 & UnrealScript & 79600 & 0.8 & 0 & Scarce \\
198 & RDoc & 79514 & 0.8 & 0 & Scarce \\
199 & ShaderLab & 77504 & 0.8 & 0 & Scarce \\
200 & KiCad Schematic & 75638 & 0.8 & 0 & Scarce \\
201 & OpenSCAD & 74708 & 0.7 & 0 & Scarce \\
202 & HAProxy & 74344 & 0.7 & 0 & Scarce \\
203 & RobotFramework & 73926 & 0.7 & 0 & Scarce \\
204 & Go Checksums & 71195 & 0.7 & 0 & Scarce \\
205 & AGS Script & 71041 & 0.7 & 0 & Scarce \\
206 & SAS & 69655 & 0.7 & 0 & Scarce \\
207 & Pure Data & 69308 & 0.7 & 0 & Scarce \\
208 & Yacc & 68492 & 0.7 & 0 & Scarce \\
209 & AngelScript & 67669 & 0.7 & 0 & Scarce \\
210 & Gemfile.lock & 67462 & 0.7 & 0 & Scarce \\
211 & Tcsh & 66589 & 0.7 & 0 & Scarce \\
212 & Vim Snippet & 65973 & 0.7 & 0 & Scarce \\
213 & HTTP & 65392 & 0.7 & 0 & Scarce \\
214 & Nginx & 64585 & 0.6 & 0 & Scarce \\
215 & Textile & 64490 & 0.6 & 0 & Scarce \\
216 & Cabal Config & 64393 & 0.6 & 0 & Scarce \\
217 & AutoHotkey & 64004 & 0.6 & 0 & Scarce \\
218 & GAP & 63667 & 0.6 & 0 & Scarce \\
219 & SMT & 63299 & 0.6 & 0 & Scarce \\
220 & HTML+EEX & 63121 & 0.6 & 0 & Scarce \\
221 & BitBake & 61541 & 0.6 & 0 & Scarce \\
222 & Cython & 60384 & 0.6 & 0 & Scarce \\
223 & PureScript & 60146 & 0.6 & 0 & Scarce \\
224 & Lex & 59656 & 0.6 & 0 & Scarce \\
225 & Max & 59510 & 0.6 & 0 & Scarce \\
226 & Reason & 57730 & 0.6 & 0 & Scarce \\
227 & G-code & 54523 & 0.5 & 0 & Scarce \\
228 & VBScript & 53900 & 0.5 & 0 & Scarce \\
229 & NWScript & 52938 & 0.5 & 0 & Scarce \\
230 & ColdFusion CFC & 49873 & 0.5 & 0 & Scarce \\
231 & Xtend & 49869 & 0.5 & 0 & Scarce \\
232 & Modelica & 49219 & 0.5 & 0 & Scarce \\
233 & Eagle & 49207 & 0.5 & 0 & Scarce \\
234 & Papyrus & 48386 & 0.5 & 0 & Scarce \\
235 & ABAP & 46774 & 0.5 & 0 & Scarce \\
236 & nesC & 46665 & 0.5 & 0 & Scarce \\
237 & SWIG & 46645 & 0.5 & 0 & Scarce \\
238 & JSONLD & 45703 & 0.5 & 0 & Scarce \\
239 & SuperCollider & 45464 & 0.5 & 0 & Scarce \\
240 & Rebol & 44687 & 0.4 & 0 & Scarce \\
241 & Agda & 43197 & 0.4 & 0 & Scarce \\
242 & EditorConfig & 42796 & 0.4 & 0 & Scarce \\
243 & XQuery & 42556 & 0.4 & 0 & Scarce \\
244 & SourcePawn & 42411 & 0.4 & 0 & Scarce \\
245 & CSON & 40598 & 0.4 & 0 & Scarce \\
246 & SELinux Policy & 40594 & 0.4 & 0 & Scarce \\
247 & Linux Kernel Module & 40477 & 0.4 & 0 & Scarce \\
248 & Pawn & 40283 & 0.4 & 0 & Scarce \\
249 & AL & 40073 & 0.4 & 0 & Scarce \\
250 & Logos & 40071 & 0.4 & 0 & Scarce \\
251 & AspectJ & 39898 & 0.4 & 0 & Scarce \\
252 & Awk & 39800 & 0.4 & 0 & Scarce \\
253 & Protocol Buffer Text Format & 39461 & 0.4 & 0 & Scarce \\
254 & AutoIt & 39112 & 0.4 & 0 & Scarce \\
255 & Vim Help File & 38710 & 0.4 & 0 & Scarce \\
256 & OpenCL & 38697 & 0.4 & 0 & Scarce \\
257 & LiveScript & 36575 & 0.4 & 0 & Scarce \\
258 & Windows Registry Entries & 36440 & 0.4 & 0 & Scarce \\
259 & COBOL & 34905 & 0.3 & 0 & Scarce \\
260 & M4Sugar & 34688 & 0.3 & 0 & Scarce \\
261 & Eiffel & 34604 & 0.3 & 0 & Scarce \\
262 & Linker Script & 34355 & 0.3 & 0 & Scarce \\
263 & BASIC & 34168 & 0.3 & 0 & Scarce \\
264 & LookML & 33854 & 0.3 & 0 & Scarce \\
265 & Latte & 33347 & 0.3 & 0 & Scarce \\
266 & Cue Sheet & 32755 & 0.3 & 0 & Scarce \\
267 & DirectX 3D File & 32664 & 0.3 & 0 & Scarce \\
268 & JetBrains MPS & 31818 & 0.3 & 0 & Scarce \\
269 & RAML & 31284 & 0.3 & 0 & Scarce \\
270 & Volt & 30361 & 0.3 & 0 & Scarce \\
271 & Graph Modeling Language & 30282 & 0.3 & 0 & Scarce \\
272 & Ren'Py & 30228 & 0.3 & 0 & Scarce \\
273 & SPARQL & 29134 & 0.3 & 0 & Scarce \\
274 & HiveQL & 28582 & 0.3 & 0 & Scarce \\
275 & AIDL & 28577 & 0.3 & 0 & Scarce \\
276 & Lean & 28334 & 0.3 & 0 & Scarce \\
277 & Common Workflow Language & 27821 & 0.3 & 0 & Scarce \\
278 & DM & 27296 & 0.3 & 0 & Scarce \\
279 & World of Warcraft Addon Data & 27220 & 0.3 & 0 & Scarce \\
280 & GN & 27184 & 0.3 & 0 & Scarce \\
281 & PostCSS & 27016 & 0.3 & 0 & Scarce \\
282 & DTrace & 27009 & 0.3 & 0 & Scarce \\
283 & kvlang & 26345 & 0.3 & 0 & Scarce \\
284 & ZenScript & 26187 & 0.3 & 0 & Scarce \\
285 & FreeBasic & 25757 & 0.3 & 0 & Scarce \\
286 & MAXScript & 25610 & 0.3 & 0 & Scarce \\
287 & CODEOWNERS & 25378 & 0.3 & 0 & Scarce \\
288 & Alpine Abuild & 24652 & 0.2 & 0 & Scarce \\
289 & Gosu & 24554 & 0.2 & 0 & Scarce \\
290 & Portugol & 24373 & 0.2 & 0 & Scarce \\
291 & PowerBuilder & 24296 & 0.2 & 0 & Scarce \\
292 & ANTLR & 24102 & 0.2 & 0 & Scarce \\
293 & Mako & 24004 & 0.2 & 0 & Scarce \\
294 & POV-Ray SDL & 23883 & 0.2 & 0 & Scarce \\
295 & X BitMap & 23690 & 0.2 & 0 & Scarce \\
296 & Glyph Bitmap Distribution Format & 23321 & 0.2 & 0 & Scarce \\
297 & ObjectScript & 23153 & 0.2 & 0 & Scarce \\
298 & CWeb & 22893 & 0.2 & 0 & Scarce \\
299 & Dhall & 22195 & 0.2 & 0 & Scarce \\
300 & Idris & 22068 & 0.2 & 0 & Scarce \\
301 & NewLisp & 21694 & 0.2 & 0 & Scarce \\
302 & Meson & 21524 & 0.2 & 0 & Scarce \\
303 & Slash & 21422 & 0.2 & 0 & Scarce \\
304 & Ballerina & 21341 & 0.2 & 0 & Scarce \\
305 & Turing & 21270 & 0.2 & 0 & Scarce \\
306 & LabVIEW & 21172 & 0.2 & 0 & Scarce \\
307 & Zig & 20797 & 0.2 & 0 & Scarce \\
308 & sed & 20692 & 0.2 & 0 & Scarce \\
309 & M4 & 20666 & 0.2 & 0 & Scarce \\
310 & Objective-J & 19987 & 0.2 & 0 & Scarce \\
311 & Pod & 19974 & 0.2 & 0 & Scarce \\
312 & Csound Document & 19932 & 0.2 & 0 & Scarce \\
313 & CodeQL & 19180 & 0.2 & 0 & Scarce \\
314 & F* & 19084 & 0.2 & 0 & Scarce \\
315 & q & 19023 & 0.2 & 0 & Scarce \\
316 & 4D & 18929 & 0.2 & 0 & Scarce \\
317 & GSC & 18087 & 0.2 & 0 & Scarce \\
318 & ASL & 18058 & 0.2 & 0 & Scarce \\
319 & Kit & 17739 & 0.2 & 0 & Scarce \\
320 & Jasmin & 17590 & 0.2 & 0 & Scarce \\
321 & Proguard & 17548 & 0.2 & 0 & Scarce \\
322 & Literate Haskell & 17195 & 0.2 & 0 & Scarce \\
323 & RPGLE & 17136 & 0.2 & 0 & Scarce \\
324 & Adobe Font Metrics & 17030 & 0.2 & 0 & Scarce \\
325 & YANG & 17016 & 0.2 & 0 & Scarce \\
326 & Stan & 16966 & 0.2 & 0 & Scarce \\
327 & NEON & 16959 & 0.2 & 0 & Scarce \\
328 & wdl & 16876 & 0.2 & 0 & Scarce \\
329 & NCL & 16626 & 0.2 & 0 & Scarce \\
330 & V & 16619 & 0.2 & 0 & Scarce \\
331 & CLIPS & 16339 & 0.2 & 0 & Scarce \\
332 & Isabelle & 16299 & 0.2 & 0 & Scarce \\
333 & Euphoria & 15967 & 0.2 & 0 & Scarce \\
334 & Thrift & 15912 & 0.2 & 0 & Scarce \\
335 & Squirrel & 15866 & 0.2 & 0 & Scarce \\
336 & OpenQASM & 15673 & 0.2 & 0 & Scarce \\
337 & HXML & 15594 & 0.2 & 0 & Scarce \\
338 & Io & 15147 & 0.2 & 0 & Scarce \\
339 & Vala & 15079 & 0.2 & 0 & Scarce \\
340 & OpenStep Property List & 14926 & 0.1 & 0 & Scarce \\
341 & Velocity Template Language & 14882 & 0.1 & 0 & Scarce \\
342 & Nextflow & 14831 & 0.1 & 0 & Scarce \\
343 & PigLatin & 14665 & 0.1 & 0 & Scarce \\
344 & Jsonnet & 14335 & 0.1 & 0 & Scarce \\
345 & Clarion & 14109 & 0.1 & 0 & Scarce \\
346 & NPM Config & 13594 & 0.1 & 0 & Scarce \\
347 & Astro & 13465 & 0.1 & 0 & Scarce \\
348 & MoonScript & 13417 & 0.1 & 0 & Scarce \\
349 & Limbo & 13261 & 0.1 & 0 & Scarce \\
350 & Cool & 13253 & 0.1 & 0 & Scarce \\
351 & Factor & 13160 & 0.1 & 0 & Scarce \\
352 & Lasso & 12983 & 0.1 & 0 & Scarce \\
353 & Clean & 12920 & 0.1 & 0 & Scarce \\
354 & ChucK & 12882 & 0.1 & 0 & Scarce \\
355 & Redirect Rules & 12856 & 0.1 & 0 & Scarce \\
356 & Pan & 12821 & 0.1 & 0 & Scarce \\
357 & DNS Zone & 12790 & 0.1 & 0 & Scarce \\
358 & Arc & 12296 & 0.1 & 0 & Scarce \\
359 & MLIR & 12290 & 0.1 & 0 & Scarce \\
360 & WebAssembly & 12278 & 0.1 & 0 & Scarce \\
361 & Mercury & 11898 & 0.1 & 0 & Scarce \\
362 & YARA & 11810 & 0.1 & 0 & Scarce \\
363 & Inno Setup & 11765 & 0.1 & 0 & Scarce \\
364 & MQL4 & 11585 & 0.1 & 0 & Scarce \\
365 & OpenType Feature File & 11411 & 0.1 & 0 & Scarce \\
366 & Promela & 11300 & 0.1 & 0 & Scarce \\
367 & Marko & 11114 & 0.1 & 0 & Scarce \\
368 & Wollok & 11078 & 0.1 & 0 & Scarce \\
369 & MQL5 & 10666 & 0.1 & 0 & Scarce \\
370 & Sage & 10633 & 0.1 & 0 & Scarce \\
371 & Modula-2 & 10551 & 0.1 & 0 & Scarce \\
372 & API Blueprint & 10522 & 0.1 & 0 & Scarce \\
373 & Easybuild & 10439 & 0.1 & 0 & Scarce \\
374 & Texinfo & 10330 & 0.1 & 0 & Scarce \\
375 & Parrot Internal Representation & 10266 & 0.1 & 0 & Scarce \\
376 & Ceylon & 10193 & 0.1 & 0 & Scarce \\
377 & Red & 10159 & 0.1 & 0 & Scarce \\
378 & Rascal & 9949 & 0.1 & 0 & Scarce \\
379 & Brainfuck & 9854 & 0.1 & 0 & Scarce \\
380 & NSIS & 9820 & 0.1 & 0 & Scarce \\
381 & Ninja & 9600 & 0.1 & 0 & Scarce \\
382 & Monkey C & 9082 & 0.1 & 0 & Scarce \\
383 & Open Policy Agent & 9028 & 0.1 & 0 & Scarce \\
384 & JSON5 & 8962 & 0.1 & 0 & Scarce \\
385 & Modula-3 & 8861 & 0.1 & 0 & Scarce \\
386 & Prisma & 8835 & 0.1 & 0 & Scarce \\
387 & Literate CoffeeScript & 8619 & 0.1 & 0 & Scarce \\
388 & BlitzMax & 8557 & 0.1 & 0 & Scarce \\
389 & Debian Package Control File & 8552 & 0.1 & 0 & Scarce \\
390 & ImageJ Macro & 8551 & 0.1 & 0 & Scarce \\
391 & Readline Config & 8308 & 0.1 & 0 & Scarce \\
392 & Bicep & 8274 & 0.1 & 0 & Scarce \\
393 & Boo & 8238 & 0.1 & 0 & Scarce \\
394 & GAMS & 8195 & 0.1 & 0 & Scarce \\
395 & Csound Score & 8182 & 0.1 & 0 & Scarce \\
396 & JFlex & 8105 & 0.1 & 0 & Scarce \\
397 & Genero & 8084 & 0.1 & 0 & Scarce \\
398 & Microsoft Developer Studio Project & 8080 & 0.1 & 0 & Scarce \\
399 & Genero Forms & 8053 & 0.1 & 0 & Scarce \\
400 & J & 8032 & 0.1 & 0 & Scarce \\
401 & Xojo & 7973 & 0.1 & 0 & Scarce \\
402 & CUE & 7867 & 0.1 & 0 & Scarce \\
403 & GDB & 7784 & 0.1 & 0 & Scarce \\
404 & Alloy & 7764 & 0.1 & 0 & Scarce \\
405 & Bluespec & 7624 & 0.1 & 0 & Scarce \\
406 & BlitzBasic & 7612 & 0.1 & 0 & Scarce \\
407 & Metal & 7483 & 0.1 & 0 & Scarce \\
408 & P4 & 7453 & 0.1 & 0 & Scarce \\
409 & TLA & 7303 & 0.1 & 0 & Scarce \\
410 & Ring & 7299 & 0.1 & 0 & Scarce \\
411 & STAR & 7210 & 0.1 & 0 & Scarce \\
412 & Monkey & 7192 & 0.1 & 0 & Scarce \\
413 & Boogie & 7145 & 0.1 & 0 & Scarce \\
414 & RPC & 7024 & 0.1 & 0 & Scarce \\
415 & NASL & 7003 & 0.1 & 0 & Scarce \\
416 & Macaulay2 & 6970 & 0.1 & 0 & Scarce \\
417 & Pony & 6928 & 0.1 & 0 & Scarce \\
418 & Oz & 6774 & 0.1 & 0 & Scarce \\
419 & LTspice Symbol & 6717 & 0.1 & 0 & Scarce \\
420 & Singularity & 6669 & 0.1 & 0 & Scarce \\
421 & Browserslist & 6634 & 0.1 & 0 & Scarce \\
422 & ATS & 6633 & 0.1 & 0 & Scarce \\
423 & Fantom & 6622 & 0.1 & 0 & Scarce \\
424 & Genshi & 6608 & 0.1 & 0 & Scarce \\
425 & APL & 6463 & 0.1 & 0 & Scarce \\
426 & Berry & 6463 & 0.1 & 0 & Scarce \\
427 & Dafny & 6224 & 0.1 & 0 & Scarce \\
428 & SSH Config & 6136 & 0.1 & 0 & Scarce \\
429 & Terra & 6090 & 0.1 & 0 & Scarce \\
430 & Nemerle & 6064 & 0.1 & 0 & Scarce \\
431 & LSL & 6056 & 0.1 & 0 & Scarce \\
432 & ASN.1 & 6013 & 0.1 & 0 & Scarce \\
433 & Asymptote & 6010 & 0.1 & 0 & Scarce \\
434 & Witcher Script & 5987 & 0.1 & 0 & Scarce \\
435 & Zephir & 5890 & 0.1 & 0 & Scarce \\
436 & Uno & 5804 & 0.1 & 0 & Scarce \\
437 & 2-Dimensional Array & 5766 & 0.1 & 0 & Scarce \\
438 & RenderScript & 5751 & 0.1 & 0 & Scarce \\
439 & E & 5749 & 0.1 & 0 & Scarce \\
440 & ECL & 5748 & 0.1 & 0 & Scarce \\
441 & EmberScript & 5724 & 0.1 & 0 & Scarce \\
442 & Closure Templates & 5713 & 0.1 & 0 & Scarce \\
443 & Brightscript & 5681 & 0.1 & 0 & Scarce \\
444 & REXX & 5660 & 0.1 & 0 & Scarce \\
445 & Csound & 5653 & 0.1 & 0 & Scarce \\
446 & SugarSS & 5423 & 0.1 & 0 & Scarce \\
447 & Zeek & 5378 & 0.1 & 0 & Scarce \\
448 & Inform 7 & 5361 & 0.1 & 0 & Scarce \\
449 & Type Language & 5310 & 0.1 & 0 & Scarce \\
450 & Futhark & 4925 & 0.0 & 0 & Scarce \\
451 & XS & 4917 & 0.0 & 0 & Scarce \\
452 & CAP CDS & 4699 & 0.0 & 0 & Scarce \\
453 & DataWeave & 4682 & 0.0 & 0 & Scarce \\
454 & MiniYAML & 4669 & 0.0 & 0 & Scarce \\
455 & OpenRC runscript & 4630 & 0.0 & 0 & Scarce \\
456 & Q\# & 4615 & 0.0 & 0 & Scarce \\
457 & Propeller Spin & 4575 & 0.0 & 0 & Scarce \\
458 & Spline Font Database & 4569 & 0.0 & 0 & Scarce \\
459 & Soong & 4457 & 0.0 & 0 & Scarce \\
460 & Dylan & 4397 & 0.0 & 0 & Scarce \\
461 & XC & 4396 & 0.0 & 0 & Scarce \\
462 & CartoCSS & 4354 & 0.0 & 0 & Scarce \\
463 & Bikeshed & 4353 & 0.0 & 0 & Scarce \\
464 & Faust & 4304 & 0.0 & 0 & Scarce \\
465 & Cadence & 4295 & 0.0 & 0 & Scarce \\
466 & jq & 4274 & 0.0 & 0 & Scarce \\
467 & ObjDump & 4223 & 0.0 & 0 & Scarce \\
468 & Move & 4205 & 0.0 & 0 & Scarce \\
469 & Nearley & 4202 & 0.0 & 0 & Scarce \\
470 & X10 & 4200 & 0.0 & 0 & Scarce \\
471 & Grammatical Framework & 4193 & 0.0 & 0 & Scarce \\
472 & Altium Designer & 4167 & 0.0 & 0 & Scarce \\
473 & ooc & 4038 & 0.0 & 0 & Scarce \\
474 & IGOR Pro & 3934 & 0.0 & 0 & Scarce \\
475 & Wren & 3900 & 0.0 & 0 & Scarce \\
476 & SRecode Template & 3822 & 0.0 & 0 & Scarce \\
477 & eC & 3810 & 0.0 & 0 & Scarce \\
478 & Fennel & 3798 & 0.0 & 0 & Scarce \\
479 & Motoko & 3778 & 0.0 & 0 & Scarce \\
480 & NetLogo & 3775 & 0.0 & 0 & Scarce \\
481 & EQ & 3759 & 0.0 & 0 & Scarce \\
482 & Harbour & 3626 & 0.0 & 0 & Scarce \\
483 & VCL & 3615 & 0.0 & 0 & Scarce \\
484 & Whiley & 3580 & 0.0 & 0 & Scarce \\
485 & PEG.js & 3519 & 0.0 & 0 & Scarce \\
486 & STON & 3419 & 0.0 & 0 & Scarce \\
487 & Hy & 3344 & 0.0 & 0 & Scarce \\
488 & Beef & 3288 & 0.0 & 0 & Scarce \\
489 & StringTemplate & 3287 & 0.0 & 0 & Scarce \\
490 & LOLCODE & 3278 & 0.0 & 0 & Scarce \\
491 & hoon & 3250 & 0.0 & 0 & Scarce \\
492 & Ragel & 3232 & 0.0 & 0 & Scarce \\
493 & IRC log & 3184 & 0.0 & 0 & Scarce \\
494 & XProc & 3182 & 0.0 & 0 & Scarce \\
495 & Pic & 3154 & 0.0 & 0 & Scarce \\
496 & Odin & 3070 & 0.0 & 0 & Scarce \\
497 & mupad & 3003 & 0.0 & 0 & Scarce \\
498 & Jolie & 2941 & 0.0 & 0 & Scarce \\
499 & ABAP CDS & 2840 & 0.0 & 0 & Scarce \\
500 & Frege & 2828 & 0.0 & 0 & Scarce \\
501 & REALbasic & 2781 & 0.0 & 0 & Scarce \\
502 & Nu & 2781 & 0.0 & 0 & Scarce \\
503 & FLUX & 2775 & 0.0 & 0 & Scarce \\
504 & Pike & 2742 & 0.0 & 0 & Scarce \\
505 & CIL & 2710 & 0.0 & 0 & Scarce \\
506 & KakouneScript & 2675 & 0.0 & 0 & Scarce \\
507 & nanorc & 2673 & 0.0 & 0 & Scarce \\
508 & GEDCOM & 2669 & 0.0 & 0 & Scarce \\
509 & Literate Agda & 2654 & 0.0 & 0 & Scarce \\
510 & Cloud Firestore Security Rules & 2600 & 0.0 & 0 & Scarce \\
511 & Kusto & 2550 & 0.0 & 0 & Scarce \\
512 & C2hs Haskell & 2541 & 0.0 & 0 & Scarce \\
513 & Slice & 2468 & 0.0 & 0 & Scarce \\
514 & Filebench WML & 2465 & 0.0 & 0 & Scarce \\
515 & Component Pascal & 2465 & 0.0 & 0 & Scarce \\
516 & Chapel & 2457 & 0.0 & 0 & Scarce \\
517 & KiCad Legacy Layout & 2454 & 0.0 & 0 & Scarce \\
518 & SmPL & 2452 & 0.0 & 0 & Scarce \\
519 & MTML & 2422 & 0.0 & 0 & Scarce \\
520 & Curry & 2402 & 0.0 & 0 & Scarce \\
521 & Logtalk & 2358 & 0.0 & 0 & Scarce \\
522 & Qt Script & 2354 & 0.0 & 0 & Scarce \\
523 & Jison & 2348 & 0.0 & 0 & Scarce \\
524 & RouterOS Script & 2333 & 0.0 & 0 & Scarce \\
525 & HolyC & 2272 & 0.0 & 0 & Scarce \\
526 & EBNF & 2247 & 0.0 & 0 & Scarce \\
527 & Apollo Guidance Computer & 2222 & 0.0 & 0 & Scarce \\
528 & Tea & 2207 & 0.0 & 0 & Scarce \\
529 & SQLPL & 2204 & 0.0 & 0 & Scarce \\
530 & WebIDL & 2186 & 0.0 & 0 & Scarce \\
531 & LFE & 2144 & 0.0 & 0 & Scarce \\
532 & HTML+ECR & 2134 & 0.0 & 0 & Scarce \\
533 & KRL & 2128 & 0.0 & 0 & Scarce \\
534 & BrighterScript & 2114 & 0.0 & 0 & Scarce \\
535 & Edje Data Collection & 2101 & 0.0 & 0 & Scarce \\
536 & Creole & 2079 & 0.0 & 0 & Scarce \\
537 & GAML & 2031 & 0.0 & 0 & Scarce \\
538 & Cap'n Proto & 2027 & 0.0 & 0 & Scarce \\
539 & Zimpl & 2015 & 0.0 & 0 & Scarce \\
540 & NetLinx & 2013 & 0.0 & 0 & Scarce \\
541 & Avro IDL & 1955 & 0.0 & 0 & Scarce \\
542 & Kaitai Struct & 1896 & 0.0 & 0 & Scarce \\
543 & Janet & 1833 & 0.0 & 0 & Scarce \\
544 & UrWeb & 1818 & 0.0 & 0 & Scarce \\
545 & Talon & 1772 & 0.0 & 0 & Scarce \\
546 & PicoLisp & 1758 & 0.0 & 0 & Scarce \\
547 & MUF & 1745 & 0.0 & 0 & Scarce \\
548 & HyPhy & 1729 & 0.0 & 0 & Scarce \\
549 & Gemini & 1728 & 0.0 & 0 & Scarce \\
550 & ZIL & 1713 & 0.0 & 0 & Scarce \\
551 & DenizenScript & 1704 & 0.0 & 0 & Scarce \\
552 & Gleam & 1678 & 0.0 & 0 & Scarce \\
553 & Valve Data Format & 1672 & 0.0 & 0 & Scarce \\
554 & ECLiPSe & 1652 & 0.0 & 0 & Scarce \\
555 & Pod 6 & 1623 & 0.0 & 0 & Scarce \\
556 & Regular Expression & 1601 & 0.0 & 0 & Scarce \\
557 & JSONiq & 1589 & 0.0 & 0 & Scarce \\
558 & Muse & 1582 & 0.0 & 0 & Scarce \\
559 & Antlers & 1568 & 0.0 & 0 & Scarce \\
560 & Riot & 1537 & 0.0 & 0 & Scarce \\
561 & Mint & 1519 & 0.0 & 0 & Scarce \\
562 & XPages & 1496 & 0.0 & 0 & Scarce \\
563 & Cycript & 1482 & 0.0 & 0 & Scarce \\
564 & Pep8 & 1461 & 0.0 & 0 & Scarce \\
565 & Git Revision List & 1426 & 0.0 & 0 & Scarce \\
566 & Elvish & 1425 & 0.0 & 0 & Scarce \\
567 & Xonsh & 1386 & 0.0 & 0 & Scarce \\
568 & Click & 1358 & 0.0 & 0 & Scarce \\
569 & Nasal & 1280 & 0.0 & 0 & Scarce \\
570 & Grace & 1277 & 0.0 & 0 & Scarce \\
571 & Rouge & 1252 & 0.0 & 0 & Scarce \\
572 & NL & 1233 & 0.0 & 0 & Scarce \\
573 & Module Management System & 1229 & 0.0 & 0 & Scarce \\
574 & Gentoo Eclass & 1121 & 0.0 & 0 & Scarce \\
575 & Vyper & 1114 & 0.0 & 0 & Scarce \\
576 & Win32 Message File & 1109 & 0.0 & 0 & Scarce \\
577 & Cirru & 1075 & 0.0 & 0 & Scarce \\
578 & Unified Parallel C & 1063 & 0.0 & 0 & Scarce \\
579 & Parrot Assembly & 1062 & 0.0 & 0 & Scarce \\
580 & Lark & 1040 & 0.0 & 0 & Scarce \\
581 & Golo & 1033 & 0.0 & 0 & Scarce \\
582 & Self & 969 & 0.0 & 0 & Scarce \\
583 & dircolors & 936 & 0.0 & 0 & Scarce \\
584 & X Font Directory Index & 933 & 0.0 & 0 & Scarce \\
585 & Moocode & 927 & 0.0 & 0 & Scarce \\
586 & Redcode & 862 & 0.0 & 0 & Scarce \\
587 & ZAP & 850 & 0.0 & 0 & Scarce \\
588 & Befunge & 846 & 0.0 & 0 & Scarce \\
589 & Quake & 838 & 0.0 & 0 & Scarce \\
590 & Mirah & 813 & 0.0 & 0 & Scarce \\
591 & Ox & 789 & 0.0 & 0 & Scarce \\
592 & Nit & 784 & 0.0 & 0 & Scarce \\
593 & Fancy & 761 & 0.0 & 0 & Scarce \\
594 & TextMate Properties & 760 & 0.0 & 0 & Scarce \\
595 & PogoScript & 757 & 0.0 & 0 & Scarce \\
596 & HOCON & 738 & 0.0 & 0 & Scarce \\
597 & Yul & 729 & 0.0 & 0 & Scarce \\
598 & LigoLANG & 701 & 0.0 & 0 & Scarce \\
599 & LoomScript & 675 & 0.0 & 0 & Scarce \\
600 & ABNF & 637 & 0.0 & 0 & Scarce \\
601 & Opa & 635 & 0.0 & 0 & Scarce \\
602 & Ioke & 618 & 0.0 & 0 & Scarce \\
603 & Augeas & 601 & 0.0 & 0 & Scarce \\
604 & Mask & 580 & 0.0 & 0 & Scarce \\
605 & Scaml & 570 & 0.0 & 0 & Scarce \\
606 & Formatted & 561 & 0.0 & 0 & Scarce \\
607 & CameLIGO & 559 & 0.0 & 0 & Scarce \\
608 & Dogescript & 542 & 0.0 & 0 & Scarce \\
609 & GCC Machine Description & 534 & 0.0 & 0 & Scarce \\
610 & Shen & 532 & 0.0 & 0 & Scarce \\
611 & Isabelle ROOT & 531 & 0.0 & 0 & Scarce \\
612 & Cairo & 528 & 0.0 & 0 & Scarce \\
613 & RUNOFF & 527 & 0.0 & 0 & Scarce \\
614 & XCompose & 506 & 0.0 & 0 & Scarce \\
615 & cURL Config & 490 & 0.0 & 0 & Scarce \\
616 & Clarity & 458 & 0.0 & 0 & Scarce \\
617 & FIGlet Font & 453 & 0.0 & 0 & Scarce \\
618 & TXL & 452 & 0.0 & 0 & Scarce \\
619 & wisp & 370 & 0.0 & 0 & Scarce \\
620 & Wget Config & 366 & 0.0 & 0 & Scarce \\
621 & Glyph & 365 & 0.0 & 0 & Scarce \\
622 & Filterscript & 358 & 0.0 & 0 & Scarce \\
623 & Charity & 353 & 0.0 & 0 & Scarce \\
624 & Genie & 313 & 0.0 & 0 & Scarce \\
625 & Sieve & 306 & 0.0 & 0 & Scarce \\
626 & Bison & 256 & 0.0 & 0 & Scarce \\
627 & ShellCheck Config & 245 & 0.0 & 0 & Scarce \\
628 & ReasonLIGO & 233 & 0.0 & 0 & Scarce \\
629 & Earthly & 200 & 0.0 & 0 & Scarce \\
630 & Jison Lex & 188 & 0.0 & 0 & Scarce \\
631 & Opal & 164 & 0.0 & 0 & Scarce \\
632 & TI Program & 163 & 0.0 & 0 & Scarce \\
633 & Darcs Patch & 109 & 0.0 & 0 & Scarce \\
634 & ShellSession & 104 & 0.0 & 0 & Scarce \\
635 & Oxygene & 76 & 0.0 & 0 & Scarce \\
636 & Parrot & 64 & 0.0 & 0 & Scarce \\
637 & Myghty & 59 & 0.0 & 0 & Scarce \\
638 & Object Data Instance Notation & 40 & 0.0 & 0 & Scarce \\
639 & Ecere Projects & 34 & 0.0 & 0 & Scarce \\
640 & NumPy & 26 & 0.0 & 0 & Scarce \\
641 & Omgrofl & 17 & 0.0 & 0 & Scarce \\
642 & Python traceback & 9 & 0.0 & 0 & Scarce \\
643 & MiniD & 5 & 0.0 & 0 & Scarce \\
644 & NetLinx+ERB & 5 & 0.0 & 0 & Scarce \\
645 & Record Jar & 4 & 0.0 & 0 & Scarce \\
646 & C-ObjDump & 3 & 0.0 & 0 & Scarce \\
\end{longtable}
\end{center}

\end{document}